\documentclass[11pt]{article}

\usepackage[preprint]{acl}

\usepackage{times}
\usepackage{latexsym}
\usepackage{amsmath}
\usepackage{amssymb}
\usepackage{booktabs}
\usepackage{multirow}

\usepackage[T1]{fontenc}

\usepackage[utf8]{inputenc}

\usepackage{microtype}

\usepackage{inconsolata}

\usepackage{graphicx}

\title{Strengthening Target-Language Features:\\ SAE-Based Steering for Multilingual Inference}

\newcommand{\blfootnote}[1]{%
  \begingroup
  \renewcommand{\thefootnote}{}
  \footnote{#1}
  \addtocounter{footnote}{-1}
  \endgroup
}

\author{Hongsheng Wang \\
  Johns Hopkins University \\
  Baltimore / USA \\
  \texttt{hwang360@jh.edu} \And
  Philipp Koehn \\
  Johns Hopkins University \\
  Baltimore / USA \\
  \texttt{phi@jhu.edu} \\}

\begin{document}
\maketitle
\blfootnote{\textbf{Code:}
\href{https://github.com/HungsingWong/sae-language-steering}
{https://github.com/HungsingWong/sae-language-steering}}
\begin{abstract}
Multilingual large language models exhibit substantial
performance differences across languages, while existing
adaptation methods often require parameter updates and
considerable multilingual training data.
We propose an inference-time multilingual steering method
that uses pretrained sparse autoencoders to identify and
strengthen target-language-related features.
Using multilingual parallel sentences, we compare SAE
activations across languages and select a small number
of layer-specific features associated with each target
language.
These features are decoded into steering signals and
injected into the model's hidden states without
additional training.
Experiments with Gemma-3-12B-it show macro-average
accuracy improvements of 11.28 percentage points on
XCOPA, 4.11 points on XNLI, and 7.84 points on PAWS-X.
Additional evaluation with Qwen2.5-7B-Instruct yields
a macro-average improvement of 3.33 percentage points
on XCOPA.
\end{abstract}

\section{Introduction}

Large language models (LLMs) have demonstrated broad multilingual
capabilities, enabling a single model to process and reason across many
languages. However, this capability remains highly uneven: performance
is generally strongest in English and other high-resource languages,
while substantial gaps persist for lower-resource languages
\citep{ahuja2023mega}. One important source of this disparity is the
highly imbalanced language distribution of pretraining corpora, in which
English and a small number of high-resource languages account for a
disproportionate share of the available data
\citep{conneau2020unsupervised,muennighoff2023crosslingual}.
Consequently, improving multilingual performance without requiring
large amounts of additional language-specific training data remains an
important challenge.

Existing approaches improve multilingual performance through additional
training or representation alignment
\citep{muennighoff2023crosslingual,liu2025middle}, but require
parameter updates and substantial multilingual data.
Inference-time steering offers a lightweight alternative
\citep{turner2023steering}.
\citet{wang2025bridging} steer hidden representations primarily
toward English, whereas we steer toward the target language
by strengthening a small set of its associated SAE features.
Unlike prior work that uses language-related SAE features to identify
or control the language of generated text
\citep{deng2025unveiling,chou2025causal}, we use these features to
improve downstream multilingual task performance.

To this end, we propose an inference-time multilingual
steering method that selectively strengthens
target-language-associated features.
We compare SAE activations of parallel sentences
across languages and select a small number of features
associated with the target language.
These features are decoded into a steering signal
and applied at the final prompt position of the
final transformer layer before answer prediction.
The method requires no additional training or
parameter updates.

We evaluate our method with Gemma-3-12B-it on three
multilingual benchmarks covering commonsense reasoning,
natural language inference, and paraphrase identification.
Our method improves macro-average accuracy by
11.28 percentage points on XCOPA, 4.11 points on XNLI,
and 7.84 points on PAWS-X.
Additional evaluation with Qwen2.5-7B-Instruct yields
a macro-average improvement of 3.33 percentage points
on XCOPA.

Our main contributions are:
\begin{itemize}
    \item We propose an SAE-based multilingual steering method that
    identifies sparse target-language features from parallel sentences
    and strengthens them during inference without updating model
    parameters.

    \item We demonstrate improvements in macro-average
    performance on multilingual commonsense reasoning,
    natural language inference, and paraphrase identification,
    with additional cross-model evidence on XCOPA.
    
    \item Through controlled ablations, we examine the
    effects of steering-signal content, reference
    representations, steering strength, and feature budget.
    Transliteration controls further assess the sensitivity
    of the selected features to written form.
\end{itemize}

\section{Related Work}

\paragraph{Multilingual LLMs}
Although modern LLMs exhibit broad multilingual capabilities, their
performance remains uneven across languages, with particularly large
gaps between English and low-resource languages
\citep{ahuja2023mega}. This disparity is closely associated with
the highly imbalanced language distribution of pretraining corpora,
which are dominated by English and other high-resource languages. Moreover, multilingual models face a
trade-off between cross-lingual transfer and capacity dilution as the
number of supported languages increases
\citep{conneau2020unsupervised}.

\paragraph{Multilingual Adaptation and Alignment}
Multilingual capabilities are commonly improved through continued
pretraining and instruction tuning
\citep{conneau2020unsupervised,muennighoff2023crosslingual}.
Preference-based methods, including RLHF and DPO, further align model
behavior, but their multilingual extensions depend on multilingual
preference data
\citep{ouyang2022training,rafailov2023direct,dang2024rlhf}.
Representation-alignment methods such as MID-Align instead encourage
shared internal representations across languages
\citep{liu2025middle}. Although effective, these approaches require
parameter updates and substantial training data, limiting their
applicability to low-resource languages.

\paragraph{Inference-Time Representation Steering}
Early work showed that language models can be controlled by adding
steering vectors to their hidden activations
\citep{subramani2022extracting}. ActAdd later introduced an
optimization-free approach that constructs steering directions from
contrasting examples \citep{turner2023steering}.
Cross-lingual hidden states have also been shown to exhibit approximately
isomorphic word-level structures that can be aligned using orthogonal
mappings \citep{feng-etal-2025-word}.
In multilingual settings, \citet{wang2025bridging} uses a small parallel corpus to align
representations of lower-performing languages with those of a
higher-performing language during inference.

\paragraph{Sparse Autoencoders and Multilingual Representations}
Sparse autoencoders (SAEs) decompose model activations into a sparse set
of features, providing a more interpretable representation of internal
model states \citep{cunningham2023sparse,bricken2023towards}.
Recent multilingual studies have identified SAE features with
language-specific activation patterns
\citep{deng2025unveiling}. \citet{chou2025causal} further
show that intervening on a small number of these features can shift
generated text into a target language. These findings demonstrate that
language information can be localized and manipulated at the SAE
feature level.

\section{Methods}

We propose an inference-time steering method that strengthens target-language information in multilingual LLMs. We use sparse autoencoders (SAEs) to identify target-language-related features and decode them into a steering signal. The signal is added to the model's hidden states before answer prediction. Our method requires no model training or parameter updates.

\subsection{Sparse Autoencoders}

Sparse autoencoders (SAEs) decompose language-model activations into sparse
features \citep{cunningham2023sparse, bricken2023towards}. For the
hidden state $\mathbf{h}_{t}^{l} \in \mathbb{R}^{d}$ at token position $t$ and
transformer layer $l$, a layer-specific SAE encoder $f_l$ produces a sparse
feature representation
\begin{equation}
    \mathbf{z}_{t}^{l} = f_l(\mathbf{h}_{t}^{l}),
\end{equation}

where $\mathbf{z}_{t}^{l} \in \mathbb{R}^{m}$ and only a small subset of its
dimensions is active. The SAE decoder $g_l$ maps these features back to the
model's hidden space:
\begin{equation}
    \hat{\mathbf{h}}_{t}^{l} = g_l(\mathbf{z}_{t}^{l}).
\end{equation}

Each dimension of $\mathbf{z}_{t}^{l}$ is treated as an individual SAE feature.
We use pretrained, layer-specific SAEs with frozen parameters.

\subsection{Identifying Language-Associated Features}

\label{sec:Identifying Language-Specific Features}

Prior work has shown that a small number of SAE features exhibit language-specific activation patterns, and that intervening on these features can causally influence the language generated by an LLM \citep{deng2025unveiling, chou2025causal}. 

\begin{figure*}[t]
    \centering
    \includegraphics[width=\textwidth]
    {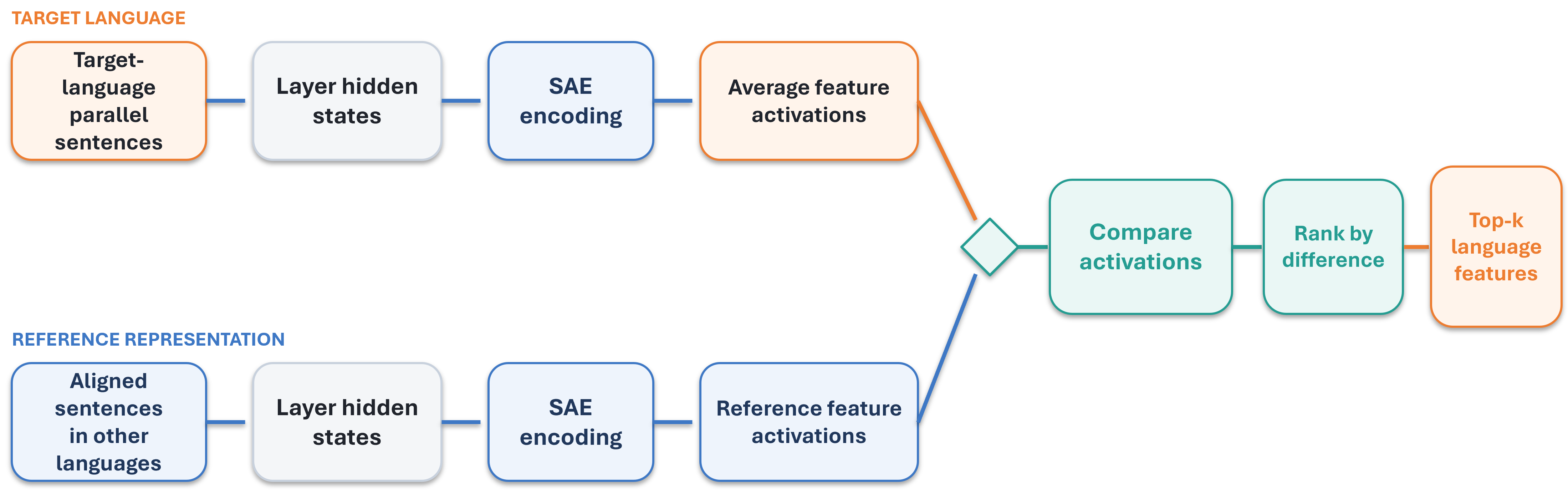}
    \caption{
    Overview of target-language feature identification.
    At each transformer layer, hidden states from aligned target- and
    reference-language sentences are encoded by an SAE and averaged
    into language-level feature activations. We rank features by the
    absolute activation difference between the target language and the
    reference representation, and select the top-$k$ features.
    The reference is either the multilingual centroid in our main
    setting or English in the comparison setting.
    }
    \label{fig:feature-identification}
\end{figure*}

Figure~\ref{fig:feature-identification} illustrates our procedure for
identifying language-discriminative SAE features at each transformer
layer.

Given a target language $\tau$, we use a multilingual parallel corpus
containing semantically aligned sentences in different languages. This
allows us to compare language-related activation patterns while
approximately controlling for semantic content.

Let $\mathbf{h}_{i,t}^{\lambda,l}$ denote the hidden state of token $t$
in the $i$-th sentence written in language $\lambda$ at layer $l$.
We encode each token-level hidden state using the SAE encoder $f^l$:
\begin{equation}
\mathbf{z}_{i,t}^{\lambda,l}
=
f^l\left(\mathbf{h}_{i,t}^{\lambda,l}\right).
\end{equation}

For each language $\lambda$, we average the SAE activations over the
tokens in each sentence and then over the $N$ parallel sentences:
\begin{equation}
\boldsymbol{\mu}_{\lambda}^{l}
=
\frac{1}{N}
\sum_{i=1}^{N}
\left(
\frac{1}{T_i^{\lambda}}
\sum_{t=1}^{T_i^{\lambda}}
\mathbf{z}_{i,t}^{\lambda,l}
\right),
\end{equation}

We consider two reference representations for identifying
target-language features. Our main method uses the average SAE
activation of all non-target languages. Following prior work on
language-specific SAE features \citep{deng2025unveiling}, we define
the set of non-target languages as
$\mathcal{L}_{-\tau}=\mathcal{L}\setminus\{\tau\}$ and compute the
multilingual centroid:
\begin{equation}
\boldsymbol{\mu}_{\mathrm{multi}}^{l}
=
\frac{1}{|\mathcal{L}_{-\tau}|}
\sum_{\lambda \in \mathcal{L}_{-\tau}}
\boldsymbol{\mu}_{\lambda}^{l}.
\end{equation}

Because the sentences are semantically aligned across languages, we
treat this multilingual centroid as an approximation of their shared,
language-neutral semantic representation. Averaging across languages
is expected to attenuate language-specific variation while preserving
information shared by the parallel sentences
\citep{kirtane2026language, sterz2025recover}.
The target-language activation difference is therefore defined as
\begin{equation}
\boldsymbol{\Delta}_{\tau,\mathrm{multi}}^{l}
=
\boldsymbol{\mu}_{\tau}^{l}
-
\boldsymbol{\mu}_{\mathrm{multi}}^{l}.
\end{equation}

\paragraph{English-reference features.}
We additionally use English as a single-language reference,
motivated by its use in prior cross-lingual steering.
This comparison examines how reference choice affects
target-language steering.
We define the target--English activation difference as
\begin{equation}
\boldsymbol{\Delta}_{\tau,\mathrm{en}}^{l}
=
\boldsymbol{\mu}_{\tau}^{l}
-
\boldsymbol{\mu}_{\mathrm{en}}^{l}.
\end{equation}

\paragraph{Feature selection.}
For either reference choice
$r \in \{\mathrm{multi}, \mathrm{en}\}$, we rank the SAE features by
the absolute magnitude of their target--reference activation
difference. At each layer $l$, we select the indices of the top-$k$
features:
\begin{equation}
\mathcal{I}_{\tau,r}^{l}
=
\operatorname{TopK}
\left(
\left|
\boldsymbol{\Delta}_{\tau,r}^{l}
\right|,
k
\right).
\end{equation}

These features exhibit the strongest activation differences between
the target language and the reference representation, and are therefore
treated as the most language-discriminative components at that layer.

\subsection{SAE Feature Steering}

\label{sec:SAE_features}

Activation steering intervenes in a model's intermediate computation by
adding a steering vector to a hidden state during inference, without
updating the model parameters
\citep{turner2023steering,zou2023representation}.
For a hidden state $\boldsymbol{h}_{q}^{l}$ at layer $l$ and token
position $q$, the intervention is generally written as
\begin{equation}
\widetilde{\boldsymbol{h}}_{q}^{l}
=
\boldsymbol{h}_{q}^{l}
+
\alpha \boldsymbol{v}^{l},
\label{eq:steering-intervention}
\end{equation}

where $\boldsymbol{v}^{l}$ is the steering vector and $\alpha$ controls
the intervention strength.

In our method, the steering vector is constructed from the SAE features
identified in Section~\ref{sec:Identifying Language-Specific Features}. For clarity, let
$\boldsymbol{\Delta}^{l}$ denote the activation difference and
$\mathcal{I}^{l}$ the selected top-$k$ feature indices at layer $l$.
We construct a sparse SAE code $\boldsymbol{s}^{l}$ as
\begin{equation}
s_j^{l}
=
\begin{cases}
\Delta_j^{l}, & j \in \mathcal{I}^{l},\\
0, & \text{otherwise}.
\end{cases}
\end{equation}

Each selected feature retains its signed target--reference activation
difference. Thus, features with larger activation contrasts contribute
more strongly to the steering signal.

All unselected SAE dimensions are set to zero.
We construct the steering vector using the SAE decoder:
\begin{equation}
v^l = -g^l(-s^l).
\end{equation}

where $g_l$ is the SAE decoder associated with layer $l$.
The resulting vector is used in the activation-steering intervention
defined above.

We apply the steering vector only at the final prompt position
of the selected transformer layer $l$, before predicting
the first response token.
Subsequent token positions are not intervened on.

\section{Experiments}

\subsection{Model and SAE}

We conduct our main experiments and ablation studies using
Gemma-3-12B-it \citep{gemmateam2025gemma3technicalreport},
an instruction-tuned 12-billion-parameter model with
48 transformer layers.
We use model-matched Gemma Scope 2 SAEs
\citep{mcdougall2025gemmascope2}, trained on the residual stream
at the output of each layer.
Each SAE contains 16,384 latent features, and we use
the small-$L_0$ variant consistently across Gemma experiments.

For cross-model evaluation, we use Qwen2.5-7B-Instruct
\citep{qwen2025qwen25technicalreport} with a pretrained BatchTopK SAE
\citep{bussmann2024batchtopksparseautoencoders}.
This SAE contains 131,072 latent features and was trained
with a sparsity budget of 32.
For both models, we select three language-associated SAE features
and keep all language-model and SAE parameters frozen
during feature identification and inference.

\subsection{Datasets}

\paragraph{Feature-identification corpus.}
We identify language-related SAE features using 500 parallel sentence
groups from FLORES-200 \citep{costa2022no}. Each group contains translations
of the same sentence across the languages considered in our experiments,
allowing us to compare feature activations while approximately controlling
for semantic content. We use these sentences only to estimate the
layer-wise SAE feature statistics described in Section~\ref{sec:Identifying Language-Specific Features}.
No downstream evaluation examples are used during feature identification.

\paragraph{Downstream evaluation.}
We evaluate the proposed intervention on three multilingual benchmarks.
XCOPA \citep{ponti2020xcopa} evaluates causal commonsense reasoning by
requiring the model to select the more plausible of two alternatives.
For each of nine target languages, we randomly split 500 examples into
100 validation examples and 400 test examples:
Estonian, Indonesian, Italian, Swahili, Tamil, Thai, Turkish, Vietnamese,
and Chinese.

XNLI \citep{conneau2018xnli} evaluates natural language inference by
classifying a premise--hypothesis pair as entailment, neutral, or
contradiction. For each of nine languages, we randomly split 500 examples into
100 validation examples and 400 test examples:
German, Spanish, French, Russian, Swahili, Thai, Turkish, Vietnamese,
and Chinese.

Finally, PAWS-X \citep{yang-etal-2019-paws} evaluates
multilingual paraphrase identification by requiring the
model to determine whether two sentences have the same meaning.
We evaluate German, Spanish, French, Japanese, and Chinese,
using the complete official validation and test splits,
each containing 2,000 examples per language.

\subsection{Language-Feature Selection}

\label{language-feature}
\begin{figure*}[t]
    \centering
    \includegraphics[width=\textwidth]
    {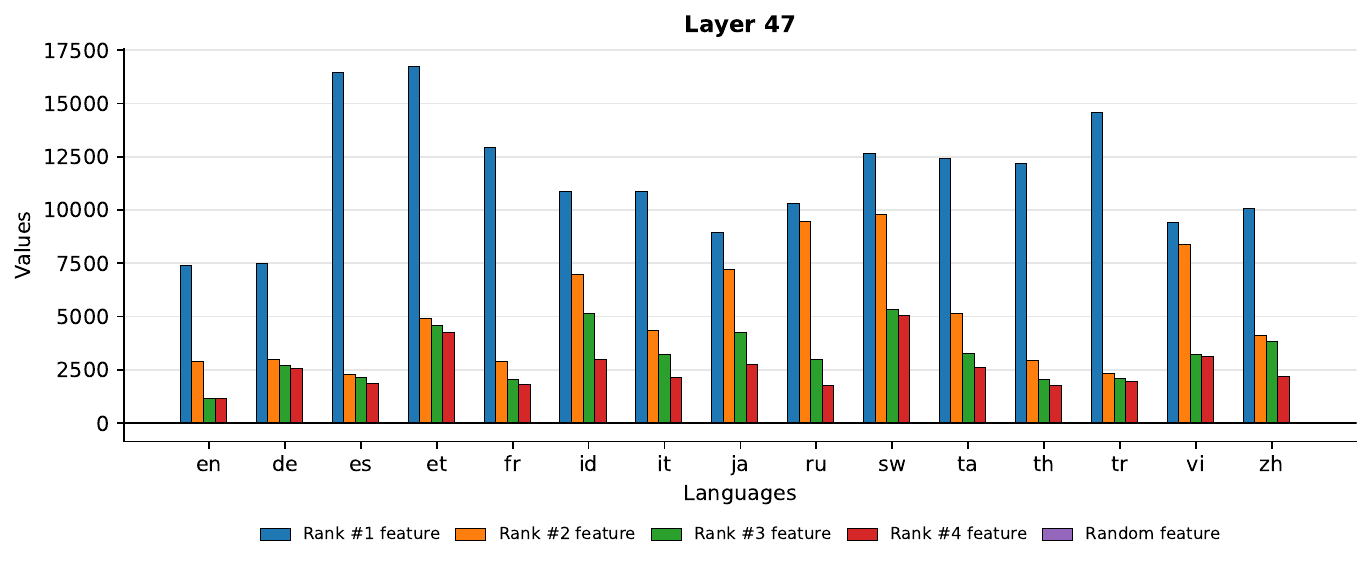}
    \caption{Target-language SAE activation differences relative to the
    multilingual centroid at layer 47. For each language, we show its four
    highest-ranked positive features and a randomly selected control
    feature. Feature identities may differ across languages.}
    \label{fig:language-features-layer30}
\end{figure*}

Figure~\ref{fig:language-features-layer30} shows the four
highest-ranked positive SAE activation differences at layer 47
for each language.
These differences substantially exceed those of the control
feature, consistent with language-associated activation patterns.
Our main intervention selects the three features with the
largest absolute activation differences, retaining both
positive and negative contrasts.

\subsection{Intervention layer.}

We evaluate every transformer layer independently on
the XCOPA validation set.
Layer 6 obtains the highest macro-average gain of
13.9 percentage points, while layer 47 achieves
a comparable gain of 13.4 points.
We retain layer 47 for the Gemma main experiments
and adopt the final transformer layer as a common
intervention location across model backbones.
This convention facilitates applying the steering
procedure to models with different depths by providing
a consistent layer-selection rule, corresponding
to layer 47 in Gemma-3-12B-it and layer 27 in
Qwen2.5-7B-Instruct.
Complete layer-wise results are provided in
Appendix~\ref{app:layer-sensitivity}.

\subsection{Evaluation and Implementation Details}

For XCOPA, XNLI, and PAWS-X, we compute the next-token
scores of the candidate labels and select the label
with the highest score.
XCOPA and PAWS-X contain two candidate labels,
while XNLI contains three.
We report accuracy for all datasets.

For each language, we select the steering coefficient $\alpha$ using
only the validation set. We search values from 0.0 to 2.5 for XCOPA and from
0.0 to 1.0 for XNLI and PAWS-X, with an interval of 0.1.
Here, $\alpha=0$ denotes the non-steered baseline. When multiple values
obtain the same tuning accuracy, we select the smallest value.

After selecting $\alpha$, we fix it and evaluate the corresponding
steered model on the test set. We also evaluate $\alpha=0$ on
the same held-out examples as the baseline. 
All experiments are conducted on three NVIDIA A100 80GB GPUs.

The complete prompt templates are provided in Appendix~\ref{app:prompts}.

\section{Results}

\paragraph{Overall performance.}
Table~\ref{tab:main-results} reports held-out test performance
for the non-steered baseline and our target-language steering
method on XCOPA, XNLI, and PAWS-X using Gemma-3-12B-it.
For each language, we select $\alpha$ on the validation split
and fix it for test evaluation.
Macro-average accuracy increases from 54.50 to 65.78 on XCOPA,
from 45.64 to 49.75 on XNLI, and from 54.78 to 62.62 on PAWS-X.
These correspond to gains of 11.28, 4.11, and 7.84 percentage
points, respectively.

\paragraph{Language-level results.}
On XCOPA, all nine evaluated languages have positive gain
estimates. The largest gains occur for Italian, Indonesian,
and Turkish, at 17.50, 15.50, and 13.75 percentage points,
respectively. Estonian shows the smallest gain at 1.50 points.

On XNLI, six languages improve over the baseline, two remain
unchanged, and German has a negative gain estimate of
2.50 points. Spanish, French, and Chinese obtain gains of
11.75, 11.25, and 9.00 points, respectively.
On PAWS-X, all five evaluated languages improve:
German gains 12.90 points, Chinese 9.70, French 7.85,
Spanish 4.50, and Japanese 4.25.
The contrasting outcomes for German on XNLI and PAWS-X
illustrate that steering benefits vary across tasks,
rather than depending on the language alone.

\paragraph{Statistical uncertainty.}
Paired bootstrap 95\% confidence intervals for the
macro-average gains are [8.05, 14.47] on XCOPA,
[1.72, 6.42] on XNLI, and [6.67, 8.99] on PAWS-X,
in percentage points.
All three intervals exclude zero, supporting positive
average gains under the evaluated settings.
The positive macro-average gains are accompanied by
task- and language-dependent variation.
For Estonian on XCOPA and German on XNLI, the observed
changes are small relative to their sampling uncertainty,
with confidence intervals spanning zero.
Complete intervals and the estimation procedure are
provided in Appendix~\ref{app:confidence-intervals}.

\begin{figure}[t]
    \centering
    \includegraphics[
        width=\columnwidth
    ]{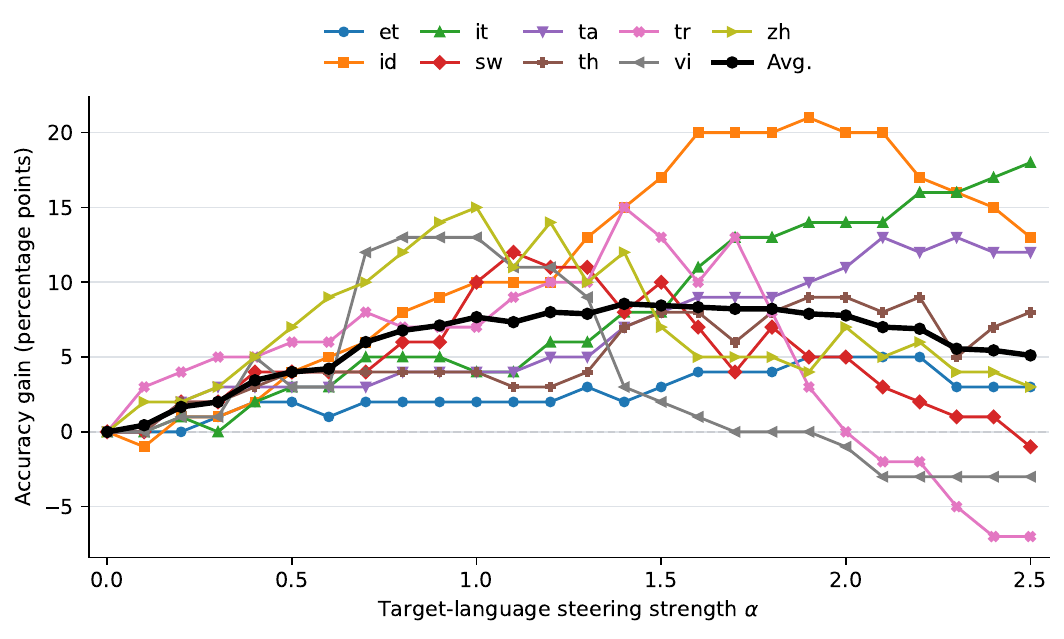}
    \caption{
        Validation accuracy gain over the non-steered baseline on XCOPA as a
        function of target-language steering strength $\alpha$.
        Colored curves correspond to individual languages, and the
        black curve reports their average.
    }
    \label{fig:alpha-xcopa}
\end{figure}

\paragraph{Effect of steering strength.}
Figure~\ref{fig:alpha-xcopa} reports the validation accuracy gain over
the non-steered baseline across the complete range of evaluated
$\alpha$ values on XCOPA. Each colored curve represents one language,
while the black curve shows the average across all nine languages.

The average gain generally increases with the target-language steering
strength and reaches its maximum of approximately $8.6$ percentage
points at $\alpha=1.4$. Performance then gradually declines as the
intervention becomes stronger. Although the optimal strength varies
across languages, most languages remain above the baseline over a broad
range of $\alpha$ values. Corresponding sensitivity results for XNLI and PAWS-X
are provided in Appendix~\ref{app:alpha-sensitivity}.

\paragraph{Performance with a shared steering coefficient.}
Using a shared coefficient of $\alpha=0.6$, held-out test
performance improves by 4.75, 1.61, and 5.13 percentage points
on XCOPA, XNLI, and PAWS-X, respectively.
The average gain remains positive on all three datasets
without language-specific coefficient selection,
although individual languages do not uniformly benefit.
Complete per-language results are provided in
Appendix~\ref{app:fixed-alpha}.

\paragraph{Cross-model evaluation.}
To examine whether the steering effects extend beyond Gemma,
we additionally evaluate Qwen2.5-7B-Instruct on the same
nine XCOPA languages, using a pretrained BatchTopK SAE
to identify the top three target-language-associated features
and apply steering at the final transformer layer.
We use the same prompt template and language-feature selection
procedure as in the Gemma experiments.
For each language, the steering strength is selected on the
validation split and fixed for held-out test evaluation.
As shown in Table~\ref{tab:qwen-xcopa}, average accuracy increases
from 58.83\% to 62.17\%, a gain of 3.33 percentage points.
Eight languages improve, while Vietnamese remains unchanged.
These results provide evidence that the proposed procedure
can yield positive steering effects with another model and
SAE family, although the gains vary across models and languages.

\begin{table*}[t]
\centering
\scriptsize
\setlength{\tabcolsep}{4pt}
\begin{tabular}{ll*{15}{c}}
\toprule
Dataset & Setting
& de & es & et & fr & id & it & ja & ru & sw & ta & th & tr & vi & zh
& Avg. \\
\midrule

\multirow{4}{*}{XCOPA}
& Baseline
& -- & -- & 54.3 & -- & 51.5 & 55.8 & -- & -- & 51.5
& 56.0 & 53.3 & 56.5 & 52.8 & 59.0 & 54.5 \\
& Ours
& -- & -- & 55.8 & -- & 67.0 & 73.3 & -- & -- & 61.0
& 66.0 & 66.8 & 70.3 & 61.5 & 70.5 & \textbf{65.8} \\
& Gain
& -- & -- & +1.5 & -- & +15.5 & +17.5 & -- & -- & +9.5
& +10.0 & +13.5 & +13.8 & +8.8 & +11.5 & \textbf{+11.3} \\
& $\alpha^\star$
& -- & -- & 1.9 & -- & 1.9 & 2.5 & -- & -- & 1.1
& 2.1 & 1.9 & 1.4 & 0.8 & 1.0 & -- \\
\midrule

\multirow{4}{*}{XNLI}
& Baseline
& 46.5 & 45.0 & -- & 44.3 & -- & -- & -- & 46.3 & 42.3
& -- & 46.0 & 47.5 & 45.8 & 47.3 & 45.6 \\
& Ours
& 44.0 & 56.8 & -- & 55.5 & -- & -- & -- & 50.5 & 44.0
& -- & 46.0 & 47.5 & 47.3 & 56.3 & \textbf{49.8} \\
& Gain
& -2.5 & +11.8 & -- & +11.3 & -- & -- & -- & +4.3 & +1.8
& -- & +0.0 & +0.0 & +1.5 & +9.0 & \textbf{+4.1} \\
& $\alpha^\star$
& 0.5 & 1.0 & -- & 0.6 & -- & -- & -- & 0.2 & 0.2
& -- & 0.2 & 0.0 & 0.1 & 0.4 & -- \\
\midrule

\multirow{4}{*}{PAWS-X}
& Baseline
& 55.85 & 55.60 & -- & 54.75 & -- & -- & 50.65 & --
& -- & -- & -- & -- & -- & 57.05 & 54.78 \\
& Ours
& 68.75 & 60.10 & -- & 62.60 & -- & -- & 54.90 & --
& -- & -- & -- & -- & -- & 66.75 & \textbf{62.62} \\
& Gain
& +12.90 & +4.50 & -- & +7.85 & -- & -- & +4.25 & --
& -- & -- & -- & -- & -- & +9.70 & \textbf{+7.84} \\
& $\alpha^\star$
& 1.0 & 1.0 & -- & 1.0 & -- & -- & 1.0 & --
& -- & -- & -- & -- & -- & 0.9 & -- \\
\bottomrule
\end{tabular}

\caption{Held-out test accuracy (\%) across XCOPA, XNLI,
and PAWS-X. Gains are reported in percentage points.
For each language, $\alpha^\star$ is selected on the
validation split and fixed for held-out test evaluation.
A dash indicates that the language is not evaluated
for the corresponding dataset. Averages and gains are computed before rounding.}
\label{tab:main-results}
\end{table*}

\begin{table}[t]
\centering
\small
\setlength{\tabcolsep}{4pt}
\begin{tabular}{lrrr}
\toprule
Language & Baseline & Ours & $\Delta$ (pp) \\
\midrule
Estonian   & 54.00 & 54.50 & +0.50 \\
Indonesian & 65.00 & 67.25 & +2.25 \\
Italian    & 65.00 & 71.50 & +6.50 \\
Swahili    & 51.50 & 52.75 & +1.25 \\
Tamil      & 51.75 & 53.00 & +1.25 \\
Thai       & 57.75 & 59.75 & +2.00 \\
Turkish    & 57.75 & 62.25 & +4.50 \\
Vietnamese & 61.50 & 61.50 & 0.00 \\
Chinese    & 65.25 & 77.00 & +11.75 \\
\midrule
Average & \textbf{58.83} & \textbf{62.17} & \textbf{+3.33} \\
\bottomrule
\end{tabular}
\caption{XCOPA test accuracy (\%) for Qwen2.5-7B-Instruct.
Steering strengths are selected separately for each language
on the validation split.
$\Delta$ denotes the gain in percentage points.
Averages and gains are computed before rounding.}
\label{tab:qwen-xcopa}
\end{table}

\section{Ablation Study}

\paragraph{SAE intervention variants.}
We compare four SAE-based variants along two axes:
the reference representation (multilingual centroid or English)
and the information included in the steering signal.
Language-only variants decode a sparse code containing only
the selected signed target--reference activation differences.
Full-representation variants apply the same feature edits
to the SAE encoding of the test hidden state and decode
the complete edited representation.

\paragraph{Direct hidden-dimension steering.}
As a non-SAE baseline, we repeat the feature-identification procedure
from Section~\ref{sec:Identifying Language-Specific Features} directly
in the residual hidden space. We select the $k$ hidden dimensions with
the largest target--multilingual activation differences and construct a
sparse steering vector from their signed values. All other settings are
kept identical to our main method. This baseline compares the
effectiveness of steering based on SAE features and direct hidden dimensions.

\paragraph{Hidden-state scaling.}
We first include a simple scaling control that does not use an SAE or a
cross-lingual reference. It directly uses the original test hidden state
as the steering signal:
\begin{equation}
\boldsymbol{v}_{\mathrm{scale}}^{l}
=
\boldsymbol{h}_{q}^{l}.
\end{equation}

Substituting this signal into the steering intervention changes the
magnitude of the original hidden state without selectively modifying
its language-related components. This control tests whether the
performance change can be explained by activation scaling alone.

\paragraph{Least-squares English projection.}
Following \citet{wang2025bridging}, we fit a linear mapping
from aligned target-language representations to English.
The projected test hidden state serves as the steering signal
at the same layer and token position as our method.
The formulation is provided in Appendix~\ref{app:ls-projection}.

Appendix Table~\ref{tab:ablation-strategies} summarizes the seven
intervention strategies and their differences.

\paragraph{Feature budget and contributions.}
We examine feature-budget sensitivity and
leave-one-feature-out effects on XCOPA.
Details and results are provided in
Appendix~\ref{app:feature-removal}
(Tables~\ref{tab:topk-sensitivity}
and~\ref{tab:xcopa-feature-removal}).

\paragraph{Transliteration control.}
To examine the sensitivity of the selected SAE features
to written form, we transliterate Chinese inputs into
pinyin on XCOPA, XNLI, and PAWS-X, and Japanese inputs
into romaji on PAWS-X.
We keep the selected features and the steering strengths
chosen on the original validation inputs fixed.
Task instructions and answer labels remain unchanged.
We compare both feature activations and task accuracy
before and after transliteration.

\begin{figure}[t]
    \centering
    \includegraphics[
        width=\columnwidth
    ]{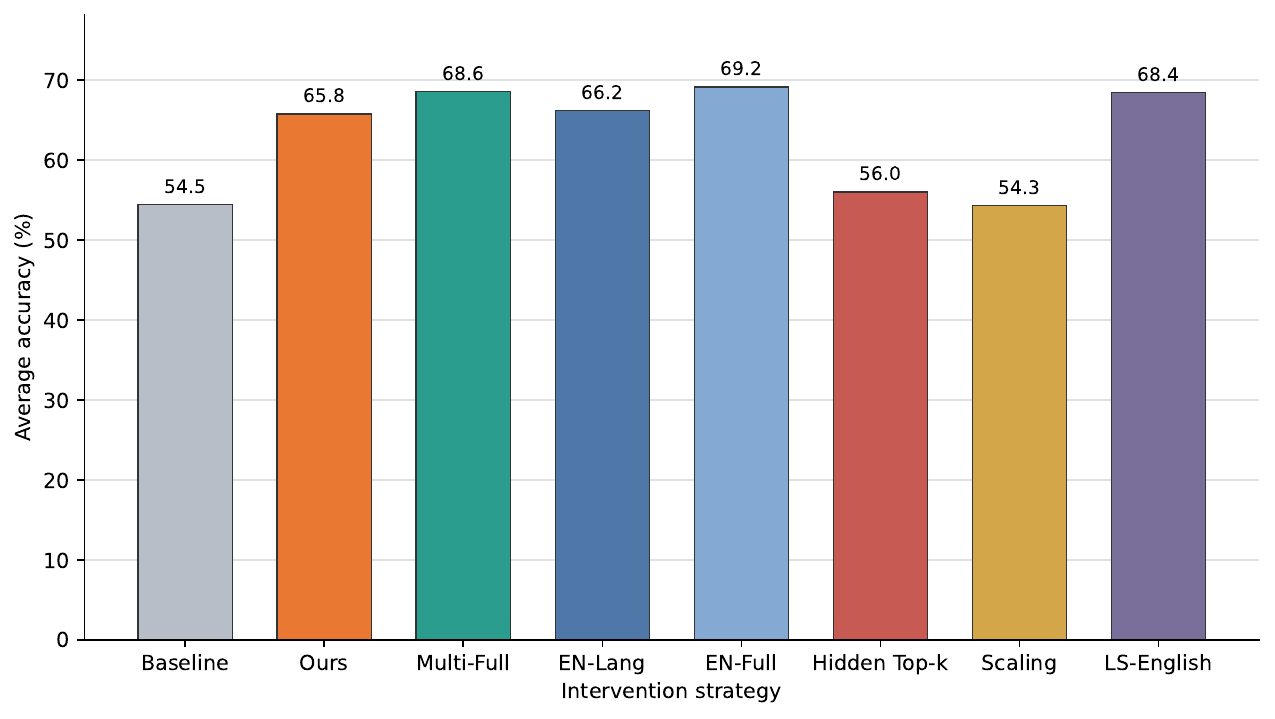}
    \caption{
        Average test accuracy of the intervention strategies on XCOPA.
        For each strategy and language, the steering strength is selected on
        the validation set before held-out evaluation. Each bar averages
        accuracy across the nine target languages.
    }
    \label{fig:xcopa-ablation}
\end{figure}

\paragraph{Ablation results.}
Figure~\ref{fig:xcopa-ablation} compares the baseline and seven
intervention strategies on XCOPA.
All four SAE-based variants outperform the baseline.
Our language-only variant reaches 65.8\%, compared with 54.5\%,
although the full-representation variants and LS-English
achieve higher accuracy.
Hidden-state scaling remains near the baseline, while direct
hidden-dimension steering yields a smaller gain.
Corresponding XNLI and PAWS-X results appear in
Appendix~\ref{app:ablation-results}.

On XCOPA, validation accuracy varies little beyond
$k=2$ or $k=3$ (Table~\ref{tab:topk-sensitivity}).
Fixing each language's validation-selected $\alpha$,
removing ranks 1, 2, and 3 reduces macro-average
test accuracy from 65.78\% to 65.53\%, 64.53\%,
and 65.39\%, respectively
(Table~\ref{tab:xcopa-feature-removal}).
The effects vary across languages: removing rank 1
increases Estonian accuracy by 10.50 percentage points
but decreases Chinese accuracy by 5.50 points.                                                                                                          

After transliteration, the selected feature activations decrease
(Table~\ref{tab:transliteration-features}).
Steering yields positive gains in three of four task--language
conditions, with Chinese XCOPA showing a gain of $-0.50$
percentage points.
Complete results and confidence intervals are provided in
Appendix~\ref{app:transliteration-control}.

\section{Analysis}
\label{sec:analysis}

\paragraph{Target-language steering improves average performance.}
Table~\ref{tab:main-results} shows that the language-only
intervention improves average performance on all three datasets.
Moreover, Figure~\ref{fig:alpha-xcopa} shows that the XCOPA
validation gain generally increases with steering strength
before declining at larger values.
These results support the effectiveness of steering with
language-associated SAE features in the evaluated settings.
Since all three benchmarks are evaluated using fixed candidate
labels, these gains cannot be explained by merely changing
the language of generated responses.

\paragraph{The effect of signal content is task-dependent.}
The language-only variants outperform the
full-representation variants on XNLI, whereas the
opposite pattern appears on XCOPA and PAWS-X.
The least-squares English projection also achieves
higher accuracy than our sparse target-language signal
on the latter two tasks.
Nevertheless, our sparse signal improves average
accuracy over the non-steered baseline on all three
main benchmarks (Table~\ref{tab:main-results}).

These comparisons show that effective steering signals
can be constructed from a small set of language-associated
SAE activation contrasts, without incorporating the
complete input representation into the signal.
Full-representation interventions can provide additional
gains, but their relative advantage depends on the task.

\paragraph{Reference effects vary across tasks.}
The multilingual language-only signal achieves higher accuracy
on XNLI, whereas the English-reference variant performs better
on XCOPA and PAWS-X.
Both variants improve average accuracy over the non-steered
baseline on all three benchmarks.
Thus, effective target-language steering is not restricted
to using English or the multilingual centroid as the reference.

\paragraph{SAE-based steering outperforms the simple controls.}
Hidden-state scaling remains close to the baseline on all
three datasets, while direct top-$k$ hidden-dimension steering
produces substantially weaker average gains than the
SAE-based interventions.
Thus, simple activation rescaling or hidden-dimension selection
does not reproduce the gains observed with SAE-based steering.

\paragraph{A compact feature budget.}
Accuracy varies little beyond $k=2$ or $k=3$ on the
XCOPA validation set (Table~\ref{tab:topk-sensitivity}).
We retain $k=3$ as a compact, near-peak configuration.

\section{Conclusion}

We introduced inference-time steering with a small set of
language-associated SAE features, without updating model parameters.
The method improves macro-average accuracy on XCOPA, XNLI,
and PAWS-X with Gemma-3-12B-it, with additional positive
evidence from Qwen2.5-7B-Instruct on XCOPA.

\section*{Limitations}

Our method relies on pretrained SAEs that are compatible
with the target model, transformer layer, and intervention point.
Although all language-model and SAE parameters remain frozen,
the applicability of the method depends on the availability
of suitable pretrained SAEs.
Our evaluation covers Gemma-3-12B-it with Gemma Scope 2
and Qwen2.5-7B-Instruct with a BatchTopK SAE.
However, the additional Qwen evaluation is limited to XCOPA,
and these results do not establish generalization across
arbitrary model architectures, SAE configurations, or tasks.

The effectiveness of steering also varies across tasks.
Compared with the main benchmarks, our Gemma experiments
on MGSM yield a substantially smaller average improvement
in multilingual mathematical reasoning.
We therefore do not claim that strengthening
language-associated features consistently improves
all downstream capabilities.
The experimental setup and per-language results for MGSM
are reported in Appendix~\ref{app:mgsm}.

\section*{Ethical Considerations}
The models, SAEs, and datasets used in this work are publicly released
and used under their respective licenses and terms of use. Our study
does not involve human participants or personal data.


\bibliography{custom}

\appendix

\section{Prompt Templates}
\label{app:prompts}

We use zero-shot prompts without in-context demonstrations. All prompts
are tokenized directly without applying an additional chat template.
The task inputs are provided in the corresponding target language,
whereas the instructions follow the English templates below.

\subsection{XCOPA}

For XCOPA, \texttt{\{relation\}} is replaced with either
\texttt{cause} or \texttt{effect}, depending on the question type.

\begin{quote}
\small\ttfamily
Here is a premise: "\{premise\}". A: "\{choice1\}" B:
"\{choice2\}" What is the \{relation\}? "A" or "B"?
\end{quote}

We compare the next-token scores assigned to \texttt{A} and \texttt{B}
and select the option with the higher score.

\subsection{XNLI}

\begin{quote}
\small\ttfamily
Premise: "[premise]"\\
Hypothesis: "[hypothesis]"\\
Which label is correct? A: entailment B: neutral C: contradiction.
Answer "A", "B", or "C"?
\end{quote}

The labels \texttt{A}, \texttt{B}, and \texttt{C} correspond to
entailment, neutral, and contradiction, respectively. We select the
label with the highest next-token score.

\subsection{PAWS-X}

\begin{quote}
\small\ttfamily
Sentence 1: "\{sentence1\}"\\
Sentence 2: "\{sentence2\}"\\
Do these two sentences have the same meaning?\\
A: yes\\
B: no\\
Answer "A" or "B"?
\end{quote}

For PAWS-X, label A denotes a paraphrase pair,
while label B denotes a non-paraphrase pair.
We compare the next-token scores of the two candidate
labels and select the label with the higher score.
The sentences remain in their original language,
while the task instructions are in English.

\section{Additional Language-Feature Analysis}
\label{app:additional-feature-analysis}

\begin{figure}[t]
    \centering
    \includegraphics[width=\columnwidth]
    {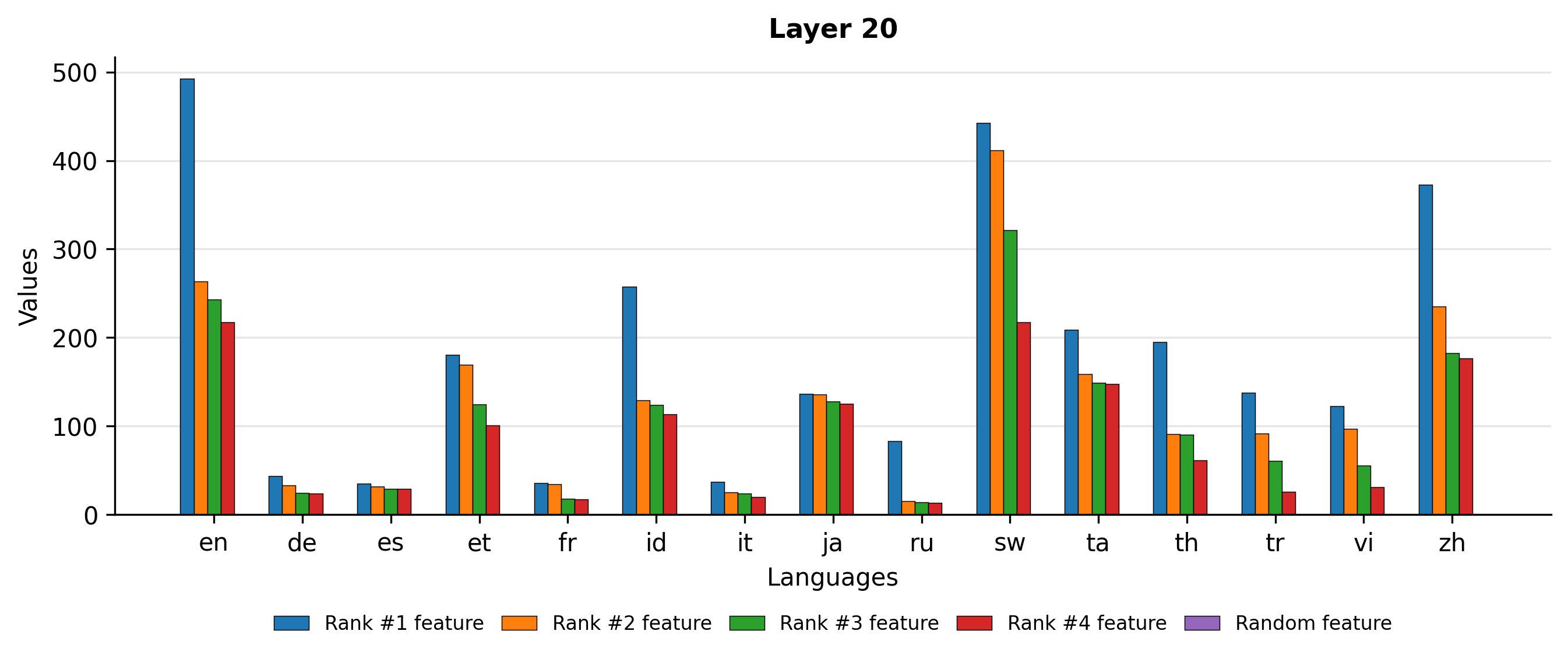}
    \caption{Target-language SAE activation differences relative to the
    multilingual centroid at layer 20.}
    \label{fig:language-features-layer20}
\end{figure}

Figure~\ref{fig:language-features-layer20} presents the corresponding
analysis at layer 20. The selected features remain substantially more
language-discriminative than the control feature, showing that the
observed concentration is not limited to layer 47.

\section{Additional Steering-Strength Results}
\label{app:alpha-sensitivity}

\begin{figure}[t]
    \centering
    \includegraphics[
        width=\columnwidth
    ]{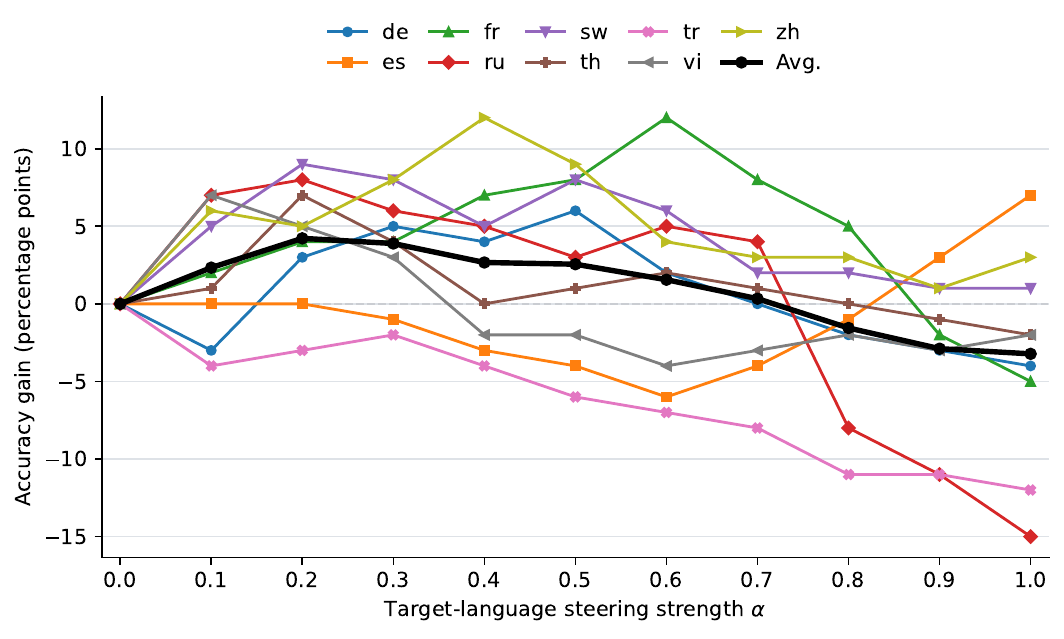}
    \caption{
        Accuracy gain over the non-steered baseline on XNLI across
        different target-language steering strengths.
    }
    \label{fig:alpha-xnli}
\end{figure}

\begin{figure}[t]
    \centering
    \includegraphics[
        width=\columnwidth
    ]{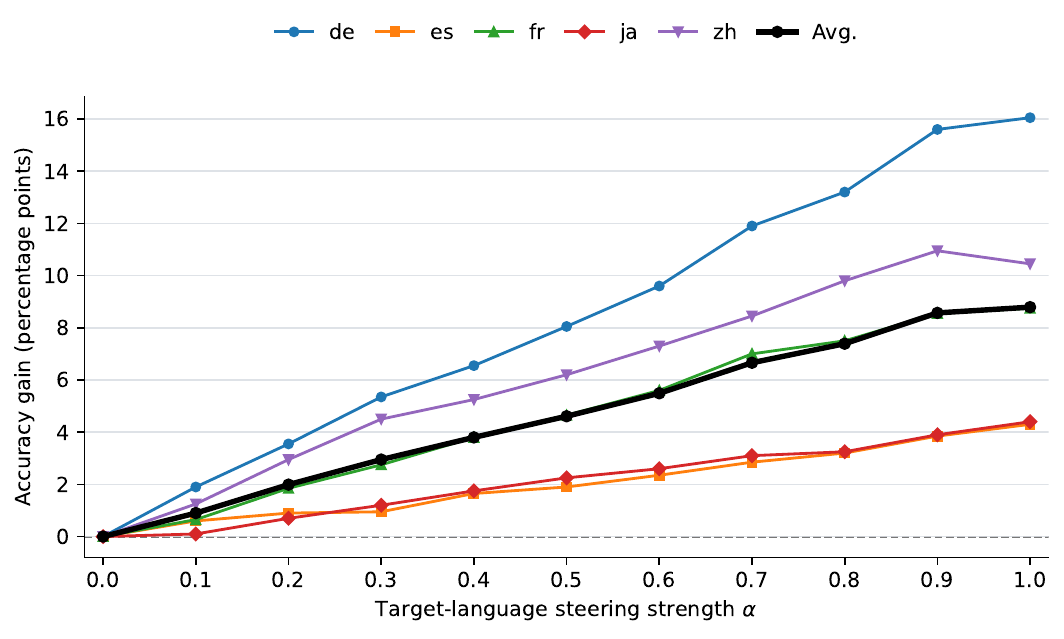}
    \caption{
        Accuracy gain over the non-steered baseline on PAWS-X across
        different target-language steering strengths.
    }
    \label{fig:alpha-pawsx}
\end{figure}

Figures~\ref{fig:alpha-xnli} and~\ref{fig:alpha-pawsx}
present the validation accuracy gains across different
steering strengths for XNLI and PAWS-X, respectively.

On XNLI, the average improvement reaches its maximum
at a relatively small steering strength, around
$\alpha=0.2$, and decreases when the intervention
becomes stronger. The individual curves also show
substantial variation across languages.

On PAWS-X, the average improvement increases across
the evaluated range of $\alpha\in[0.0,1.0]$, reaching
its highest observed value at $\alpha=1.0$.
German shows the largest gains, while Spanish and
Japanese exhibit smaller improvements.
The Chinese curve peaks at $\alpha=0.9$ and decreases
slightly at $\alpha=1.0$.

These results show that sensitivity to steering strength
varies across tasks and languages.

\section{Layer-Wise Steering Results}
\label{app:layer-sensitivity}

\begin{figure}[t]
    \centering
    \includegraphics[width=\columnwidth]
    {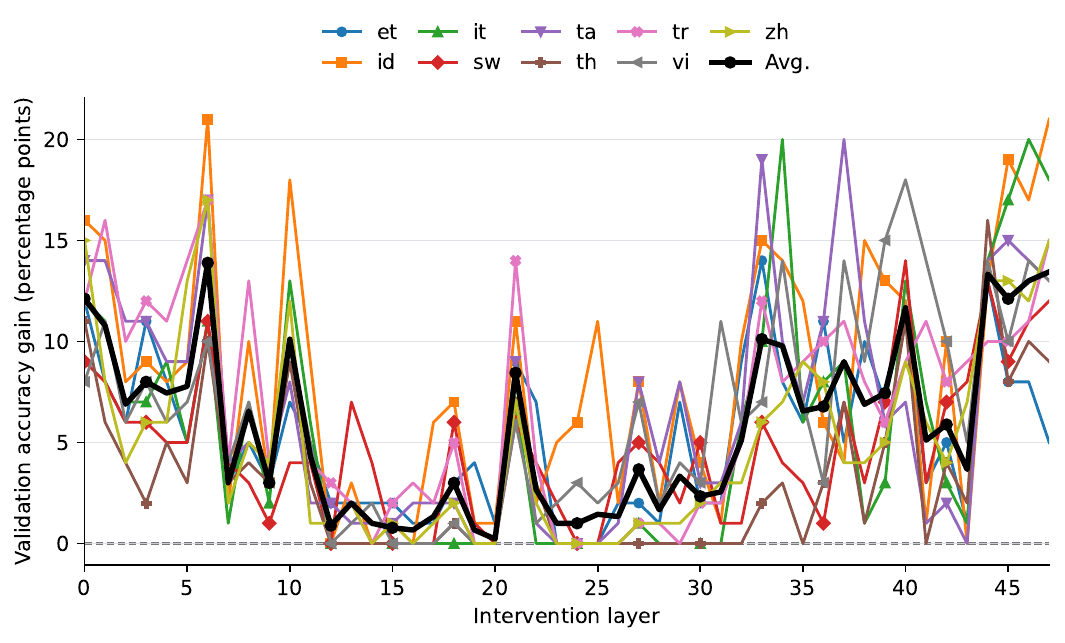}
    \caption{
    Validation accuracy gain under intervention at different transformer
    layers on XCOPA. Each layer is evaluated independently using its
    validation-selected steering strength. Colored curves denote
    individual languages, and the black curve denotes their average.
    }
    \label{fig:layer-sensitivity}
\end{figure}

Figure~\ref{fig:layer-sensitivity} reports the validation performance
obtained by intervening independently at each transformer layer. The
results show that several layers support effective steering, with
layers 6 and 47 achieving the strongest average gains.

\section{Evaluation with a Shared Steering Coefficient}
\label{app:fixed-alpha}

The main experiments select the steering coefficient using a validation
subset for each language. To examine whether the improvements depend on
language-specific coefficient selection, we conduct an additional
evaluation using a single shared coefficient. Based on the validation
trends, we set $\alpha=0.6$ and apply it unchanged to every language and
dataset. No language- or dataset-specific retuning is performed before
evaluation on the held-out test sets.

Table~\ref{tab:fixed-alpha-language} reports the complete per-language
results. The shared coefficient improves average performance by 4.75
percentage points on XCOPA, 1.61 points on XNLI, and 5.13 points on
PAWS-X. On XCOPA, eight languages improve and one remains unchanged. The
results on XNLI are more variable, with six languages improving and
three declining, while all five PAWS-X languages improve.
Nevertheless, the average gain remains positive on all three datasets,
showing that the overall effect does not require a separately optimized
coefficient for every language.

\begin{table*}[t]
\centering
\scriptsize
\setlength{\tabcolsep}{3pt}

\resizebox{\textwidth}{!}{
\begin{tabular}{ll*{15}{c}}
\toprule
Dataset & Setting
& de & es & et & fr & id & it & ja & ru
& sw & ta & th & tr & vi & zh & Avg. \\
\midrule

\multirow{3}{*}{XCOPA}
& Baseline
& -- & -- & 54.25 & -- & 51.50 & 55.75 & -- & --
& 51.50 & 56.00 & 53.25 & 56.50 & 52.75 & 59.00 & 54.50 \\
& Fixed $\alpha=0.6$
& -- & -- & 54.25 & -- & 56.75 & 59.50 & -- & --
& 56.00 & 58.75 & 56.25 & 67.00 & 59.75 & 65.00 & \textbf{59.25} \\
& Gain
& -- & -- & 0.00 & -- & +5.25 & +3.75 & -- & --
& +4.50 & +2.75 & +3.00 & +10.50 & +7.00 & +6.00 & \textbf{+4.75} \\
\midrule

\multirow{3}{*}{XNLI}
& Baseline
& 46.50 & 45.00 & -- & 44.25 & -- & -- & -- & 46.25
& 42.25 & -- & 46.00 & 47.50 & 45.75 & 47.25 & 45.64 \\
& Fixed $\alpha=0.6$
& 41.25 & 49.75 & -- & 55.50 & -- & -- & -- & 54.00
& 36.75 & -- & 46.50 & 37.25 & 46.00 & 58.25 & \textbf{47.25} \\
& Gain
& -5.25 & +4.75 & -- & +11.25 & -- & -- & -- & +7.75
& -5.50 & -- & +0.50 & -10.25 & +0.25 & +11.00 & \textbf{+1.61} \\
\midrule

\multirow{3}{*}{PAWS-X}
& Baseline
& 55.85 & 55.60 & -- & 54.75 & -- & -- & 50.65 & --
& -- & -- & -- & -- & -- & 57.05 & 54.78 \\
& Fixed $\alpha=0.6$
& 64.85 & 57.75 & -- & 59.55 & -- & -- & 52.80 & --
& -- & -- & -- & -- & -- & 64.60 & \textbf{59.91} \\
& Gain
& +9.00 & +2.15 & -- & +4.80 & -- & -- & +2.15 & --
& -- & -- & -- & -- & -- & +7.55 & \textbf{+5.13} \\

\bottomrule
\end{tabular}
}

\caption{
Per-language held-out test accuracy (\%) on XCOPA,
XNLI, and PAWS-X using a single steering coefficient
of $\alpha=0.6$.
Gains are reported in percentage points.
A dash indicates that a language is not evaluated
in the corresponding dataset.
}
\label{tab:fixed-alpha-language}
\end{table*}

\section{Least-squares English projection.}
\label{app:ls-projection}
Following the cross-lingual intervention method of \citet{wang2025bridging}, we construct a full-space transformation
toward English as a non-SAE comparison. Given aligned target-language
and English hidden states, we learn a layer-specific linear mapping
using least squares:
\begin{equation}
\boldsymbol{W}^{l}
=
\arg\min_{\boldsymbol{W}}
\left\|
\boldsymbol{H}_{\tau}^{l}\boldsymbol{W}
-
\boldsymbol{H}_{\mathrm{en}}^{l}
\right\|_{F}^{2},
\end{equation}

where $\boldsymbol{H}_{\tau}^{l}$ and
$\boldsymbol{H}_{\mathrm{en}}^{l}$ contain aligned target-language and
English representations, respectively. At inference time, we use the
projected representation
\begin{equation}
\boldsymbol{v}_{\mathrm{LS}}^{l}
=
\boldsymbol{h}_{q}^{l}\boldsymbol{W}^{l}
\end{equation}

as the steering signal. For comparability with our SAE-based method, we
apply this signal to the same residual-stream layer and at the same
intervention position. Unlike our method, this strategy transforms the
complete hidden representation rather than isolating sparse
language-related features.

\section{Intervention Strategies}
\label{app:intervention-strategies}

Table~\ref{tab:ablation-strategies} summarizes the intervention
strategies evaluated in our ablation study. The SAE-based variants
differ in their reference representation and in whether the complete
SAE encoding of the test hidden state is included in the steering signal. 
The remaining strategies serve as non-SAE controls.

\begin{table*}[t]
\centering
\renewcommand{\arraystretch}{1.2}
\setlength{\tabcolsep}{6pt}

\resizebox{\textwidth}{!}{
\begin{tabular}{lllll}
\toprule
Strategy
& Reference
& Language information
& Full test representation used?
& SAE?
\\
\midrule

Multilingual language-only (ours)
& Multilingual centroid
& Target--multilingual contrast
& No
& Yes
\\

Multilingual full representation
& Multilingual centroid
& Target--multilingual feature edit
& Yes
& Yes
\\

English language-only
& English
& Target--English contrast
& No
& Yes
\\

English full representation
& English
& Target--English feature edit
& Yes
& Yes
\\

Direct hidden-dimension steering
& Multilingual centroid
& Top-$k$ target--multilingual hidden dimensions
& No
& No
\\

Hidden-state scaling
& None
& Not isolated
& Yes
& No
\\

Least-squares English projection
& English
& Not isolated
& Yes
& No
\\

\bottomrule
\end{tabular}
}

\caption{Summary of the intervention strategies considered in the
ablation study. The SAE-based strategies vary the reference
representation and whether the complete SAE encoding of the test hidden state
is included in the steering signal. The remaining strategies serve as non-SAE controls.}
\label{tab:ablation-strategies}
\end{table*}

\section{Additional Ablation Results}
\label{app:ablation-results}

\begin{figure}[t]
    \centering
    \includegraphics[
        width=\columnwidth
    ]{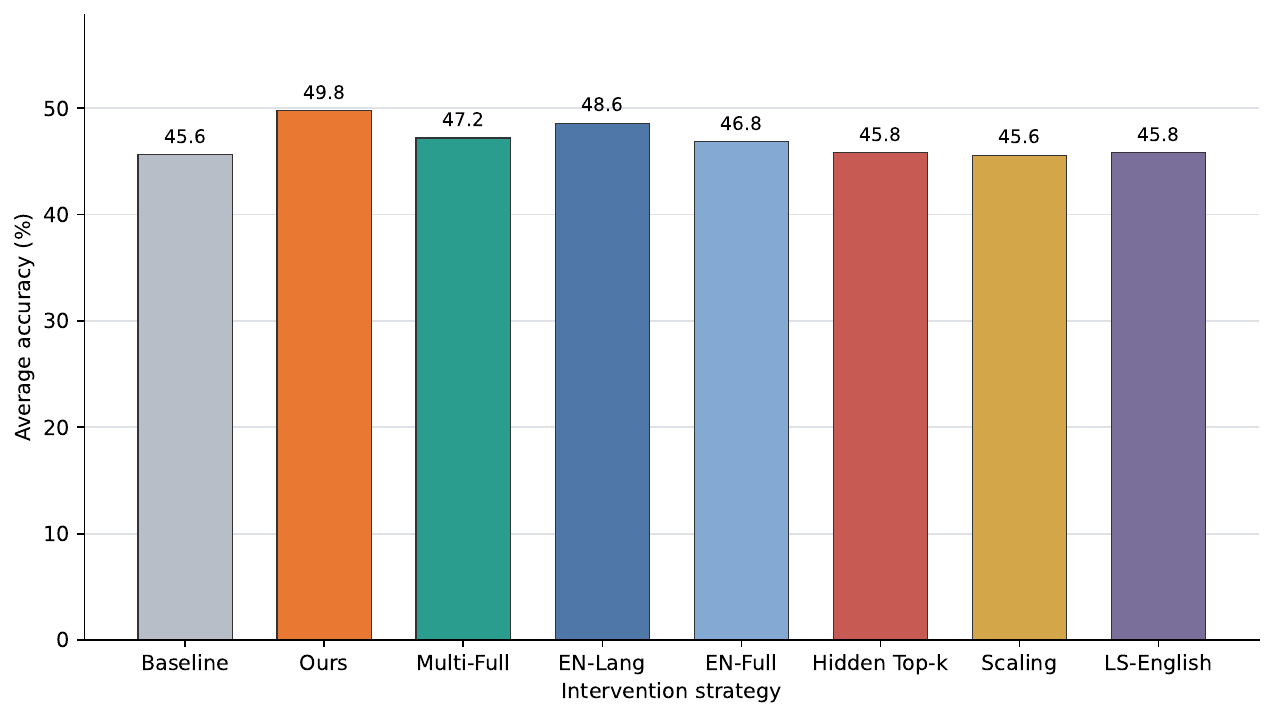}
    \caption{
        Average accuracy of the intervention strategies on XNLI.
        Each bar averages the held-out test accuracy across the nine
        target languages.
    }
    \label{fig:xnli-ablation}
\end{figure}

\paragraph{XNLI.}
Figure~\ref{fig:xnli-ablation} presents the held-out test accuracy
averaged across the nine XNLI languages. Our target-language signal
achieves the highest accuracy of 49.8, compared with the 45.6 baseline.
The English-relative language signal reaches 48.6, while the
multilingual and English full-representation variants obtain 47.2 and
46.8, respectively. Direct hidden-dimension steering and least-squares
English projection both obtain 45.8, and hidden-state scaling remains
at the baseline level with 45.6.

\begin{figure}[t]
    \centering
    \includegraphics[
        width=\columnwidth
    ]{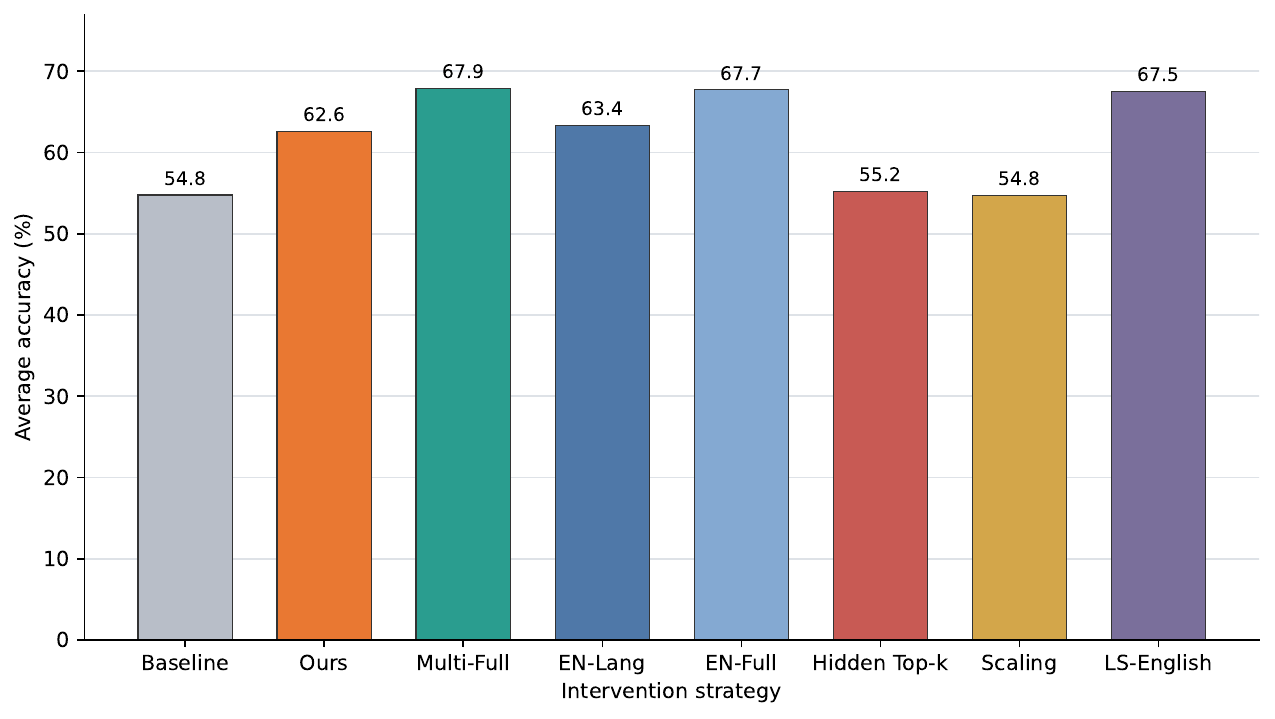}
    \caption{
        Average accuracy of the intervention strategies on PAWS-X.
        Each bar averages the held-out test accuracy across German,
        Spanish, French, Japanese, and Chinese.
    }
    \label{fig:pawsx-ablation}
\end{figure}

\paragraph{PAWS-X.}
Figure~\ref{fig:pawsx-ablation} presents the average
held-out test accuracy of the baseline and seven
intervention strategies on PAWS-X.
Our target-language signal achieves an accuracy of
62.62, compared with the 54.78 baseline.
The multilingual full-representation, English
full-representation, and least-squares English projection
strategies obtain 67.93, 67.72, and 67.53, respectively.
The English-relative language signal reaches 63.36,
while direct hidden-dimension steering obtains 55.20.
Hidden-state scaling remains close to the baseline
at 54.75.
These results show that sparse target-language steering
improves performance, although full-representation
interventions achieve larger gains on this task.

\section{Confidence Intervals}
\label{app:confidence-intervals}

For the Gemma-3-12B-it experiments, we estimate pointwise
95\% confidence intervals using paired percentile bootstrap
with 10,000 resamples and random seed 42.
We resample test examples with replacement, keeping
baseline and steered predictions paired, and compute
the accuracy difference for each resample.
The intervals use the 2.5th and 97.5th percentiles.
For macro-averages, aligned test items are resampled jointly
across languages. SAE features and validation-selected
steering strengths remain fixed throughout resampling.

\begin{table*}[t]
\centering\small
\setlength{\tabcolsep}{5pt}
\begin{tabular}{llrrrl}
\toprule
Dataset & Lang. & Baseline & Ours & $\Delta$ (pp) & 95\% CI (pp) \\
\midrule
\multirow{10}{*}{XCOPA}
& et & 54.25 & 55.75 & +1.50  & $[-0.50,\,3.50]$ \\
& id & 51.50 & 67.00 & +15.50 & $[8.00,\,22.75]$ \\
& it & 55.75 & 73.25 & +17.50 & $[13.00,\,22.00]$ \\
& sw & 51.50 & 61.00 & +9.50  & $[4.50,\,14.50]$ \\
& ta & 56.00 & 66.00 & +10.00 & $[6.75,\,13.50]$ \\
& th & 53.25 & 66.75 & +13.50 & $[9.25,\,18.00]$ \\
& tr & 56.50 & 70.25 & +13.75 & $[6.50,\,21.00]$ \\
& vi & 52.75 & 61.50 & +8.75  & $[4.25,\,13.25]$ \\
& zh & 59.00 & 70.50 & +11.50 & $[7.75,\,15.25]$ \\
& Avg. & 54.50 & 65.78 & \textbf{+11.28} & $[8.05,\,14.47]$ \\
\midrule
\multirow{10}{*}{XNLI}
& de & 46.50 & 44.00 & -2.50  & $[-8.25,\,3.00]$ \\
& es & 45.00 & 56.75 & +11.75 & $[6.25,\,17.75]$ \\
& fr & 44.25 & 55.50 & +11.25 & $[5.50,\,17.00]$ \\
& ru & 46.25 & 50.50 & +4.25  & $[1.00,\,7.50]$ \\
& sw & 42.25 & 44.00 & +1.75  & $[-2.00,\,5.50]$ \\
& th & 46.00 & 46.00 & 0.00   & $[-3.25,\,3.25]$ \\
& tr & 47.50 & 47.50 & 0.00   & $[0.00,\,0.00]$ \\
& vi & 45.75 & 47.25 & +1.50  & $[-1.25,\,4.25]$ \\
& zh & 47.25 & 56.25 & +9.00  & $[4.00,\,14.25]$ \\
& Avg. & 45.64 & 49.75 & \textbf{+4.11} & $[1.72,\,6.42]$ \\
\midrule
\multirow{6}{*}{PAWS-X}
& de & 55.85 & 68.75 & +12.90 & $[10.05,\,15.65]$ \\
& es & 55.60 & 60.10 & +4.50  & $[3.30,\,5.65]$ \\
& fr & 54.75 & 62.60 & +7.85  & $[6.20,\,9.45]$ \\
& ja & 50.65 & 54.90 & +4.25  & $[3.10,\,5.45]$ \\
& zh & 57.05 & 66.75 & +9.70  & $[7.40,\,12.00]$ \\
& Avg. & 54.78 & 62.62 & \textbf{+7.84} & $[6.67,\,8.99]$ \\
\bottomrule
\end{tabular}
\caption{Gemma-3-12B-it held-out test accuracy (\%) and
95\% confidence intervals for accuracy gains.
Each language has 400 test examples on XCOPA and XNLI,
and 2,000 on PAWS-X. Avg.\ denotes the macro-average.}
\label{tab:gain-ci}
\end{table*}

All three macro-average gain intervals exclude zero,
while language-level uncertainty varies.
The intervals for Estonian on XCOPA and German on XNLI
include zero, so these differences are not clearly resolved
by the evaluated test samples.
The zero-width interval for Turkish on XNLI reflects
the selection of $\alpha=0$, yielding identical predictions.

\section{Transliteration Control}
\label{app:transliteration-control}

We follow the main experimental protocol, converting
Chinese task content into tone-numbered pinyin and Japanese
content into Hepburn romaji using pypinyin v0.55.0 and
pykakasi v2.3.0, respectively.
The original feature sets and validation-selected steering
strengths remain fixed, without reselection or retuning.
Instructions and answer labels are unchanged.

\begin{table*}[t]
\centering\small
\begin{tabular}{llrllrl}
\toprule
Dataset & Lang. & $\alpha$ & Original & Romanized
& Gain & 95\% CI \\
\midrule
XCOPA & zh & 1.0 & $59.00\to70.50$ & $50.25\to49.75$
& -0.50 & $[-2.75,1.75]$ \\
XNLI & zh & 0.4 & $47.25\to56.25$ & $35.00\to49.75$
& +14.75 & $[8.75,20.75]$ \\
PAWS-X & zh & 0.9 & $57.05\to66.75$ & $50.45\to57.85$
& +7.40 & $[4.50,10.25]$ \\
PAWS-X & ja & 1.0 & $50.65\to54.90$ & $47.65\to52.40$
& +4.75 & $[2.50,7.00]$ \\
\bottomrule
\end{tabular}
\caption{Held-out test accuracy (\%) before and after
transliteration. Each arrow denotes baseline $\to$ steered
accuracy. Gains and 95\% confidence intervals refer to
steering on romanized inputs and are in percentage points.}
\label{tab:transliteration-results}
\end{table*}

\begin{table}[t]
\centering\small
\setlength{\tabcolsep}{4pt}
\begin{tabular}{llll}
\toprule
Dataset & Lang. & Feature IDs & Retention (\%) \\
\midrule
XCOPA & zh & 108, 509, 902 & 2.09, 39.72, 42.19 \\
XNLI & zh & 108, 509, 902 & 0.76, 26.78, 27.03 \\
PAWS-X & zh & 108, 509, 902 & 0.34, 22.40, 18.00 \\
PAWS-X & ja & 809, 228, 586 & 26.14, 1.54, 7.27 \\
\bottomrule
\end{tabular}
\caption{Feature-activation retention after transliteration,
listed in feature-ID order. Retention is the romanized-to-original
mean activation ratio, expressed as a percentage.
Activations are measured on unsteered passes and averaged
over prompt tokens within each example, then over examples.}
\label{tab:transliteration-features}
\end{table}

Transliteration also changes tokenization and input distribution,
so this control does not isolate script alone.

\section{Feature Budget and Feature Contributions}
\label{app:feature-removal}

Both analyses use Gemma-3-12B-it at layer 47 on nine
XCOPA languages. For feature removal, we remove each
of the main method's three selected features in turn,
keeping the original validation-selected $\alpha$ and
remaining feature weights fixed. Decoder bias is retained,
and all removal variants are evaluated on the held-out test split.

\begin{table}[t]
\centering
\small
\setlength{\tabcolsep}{4pt}
\begin{tabular}{lccccc}
\toprule
$k$ & 1 & 2 & 3 & 4 & 5 \\
\midrule
Accuracy (\%)
& 67.78
& \textbf{69.44}
& \textbf{69.44}
& 69.33
& 69.33 \\
\bottomrule
\end{tabular}
\caption{
Effect of the number of selected SAE features.
Scores are macro-average XCOPA validation accuracy (\%)
across nine languages using Gemma-3-12B-it at layer 47.
Steering strength is selected separately for each
language and each value of $k$.
}
\label{tab:topk-sensitivity}
\end{table}

\begin{table*}[t]
\centering
\small
\setlength{\tabcolsep}{3pt}
\resizebox{\textwidth}{!}{
\begin{tabular}{lrrrrrrr}
\toprule
Lang. & Full & w/o 1 & $C_1$ [95\% CI]
& w/o 2 & $C_2$ [95\% CI]
& w/o 3 & $C_3$ [95\% CI] \\
\midrule
et & 55.75 & 66.25 & -10.50 [-16.00, -5.00] & 54.50 & +1.25 [-0.50, +3.00] & 55.50 & +0.25 [-0.75, +1.25] \\
id & 67.00 & 69.25 & -2.25 [-6.00, +1.50] & 64.25 & +2.75 [-3.75, +9.00] & 65.50 & +1.50 [-2.25, +5.25] \\
it & 73.25 & 73.75 & -0.50 [-2.25, +1.25] & 70.25 & +3.00 [+0.75, +5.50] & 73.50 & -0.25 [-1.50, +1.00] \\
sw & 61.00 & 58.00 & +3.00 [-1.25, +7.25] & 61.25 & -0.25 [-1.25, +0.50] & 61.75 & -0.75 [-4.00, +2.50] \\
ta & 66.00 & 72.00 & -6.00 [-9.00, -3.25] & 63.00 & +3.00 [+1.25, +5.00] & 63.50 & +2.50 [+1.00, +4.25] \\
th & 66.75 & 59.50 & +7.25 [+3.75, +10.75] & 67.75 & -1.00 [-2.75, +0.75] & 65.75 & +1.00 [+0.25, +2.00] \\
tr & 70.25 & 67.75 & +2.50 [-3.50, +8.75] & 71.50 & -1.25 [-3.25, +0.75] & 70.50 & -0.25 [-1.25, +0.75] \\
vi & 61.50 & 58.25 & +3.25 [-0.50, +7.00] & 58.25 & +3.25 [-0.25, +6.75] & 63.00 & -1.50 [-4.50, +1.50] \\
zh & 70.50 & 65.00 & +5.50 [+3.00, +8.25] & 70.00 & +0.50 [+0.00, +1.25] & 69.50 & +1.00 [+0.00, +2.25] \\
\midrule
Avg. & \textbf{65.78} & 65.53 & +0.25
& 64.53 & +1.25 & 65.39 & +0.39 \\
\bottomrule
\end{tabular}
}
\caption{Leave-one-feature-out results on XCOPA.
Full uses all three selected features; w/o $r$ removes
the feature ranked $r$ by absolute activation contrast.
Accuracies are percentages. Contributions
$C_r=\mathrm{Acc}_{\mathrm{Full}}-\mathrm{Acc}_{\mathrm{w/o}\,r}$
are in percentage points, with paired bootstrap 95\%
confidence intervals. The average row reports point estimates only.}
\label{tab:xcopa-feature-removal}
\end{table*}

\section{Additional Evaluation on MGSM}
\label{app:mgsm}

\paragraph{Experimental setup.}
We evaluate multilingual mathematical reasoning on MGSM\citep{shi2022language}
using Gemma-3-12B-it with Gemma Scope 2 SAEs.
We consider German, Spanish, French, and Japanese.
For each language, the 250 examples are randomly divided
into 50 validation and 200 test examples using seed 42.
We select $\alpha$ from 0.0 to 1.0 in increments of 0.1
on the validation split and fix it for test evaluation.
We use greedy decoding with at most 600 generated tokens,
extract the final numerical answer, and report exact-match accuracy.

\paragraph{Prompt template.}
\begin{quote}
\small\ttfamily
Below is an instruction that describes a task.
Write a response that appropriately completes the request
in \{language\}. Please answer in \{language\}.\\[0.5em]
\#\#\# Instruction:\\
\{question\}\\[0.5em]
\#\#\# Response:
\end{quote}

\paragraph{Results.}
Table~\ref{tab:mgsm-results} reports the held-out test results
with validation-selected steering strengths.

\begin{table}[t]
\centering\small
\begin{tabular}{lrrrr}
\toprule
Language & Baseline & Ours & $\Delta$ (pp) & $\alpha^\star$ \\
\midrule
German   & 82.0 & 84.5 & +2.5 & 0.6 \\
Spanish  & 82.0 & 82.0 & 0.0  & 0.0 \\
French   & 77.0 & 77.5 & +0.5 & 0.1 \\
Japanese & 81.0 & 80.5 & -0.5 & 0.9 \\
\midrule
Average  & 80.5 & 81.1 & +0.6 & -- \\
\bottomrule
\end{tabular}
\caption{MGSM held-out test exact-match accuracy (\%).
$\Delta$ denotes the gain in percentage points.
$\alpha^\star$ is selected separately for each language
on the validation split.}
\label{tab:mgsm-results}
\end{table}

With a shared coefficient of $\alpha=0.6$, test accuracies
are 84.5\%, 81.5\%, 77.5\%, and 87.0\% for German,
Spanish, French, and Japanese, respectively.
The macro-average is 82.63\%, a gain of 2.13 percentage points.

\paragraph{Discussion.}
The average improvements on MGSM are smaller than those
on the main benchmarks, and the outcomes vary across
languages and steering strengths.
These results illustrate the task-dependent effectiveness
of our method and do not establish uniform improvements
in multilingual mathematical reasoning.

\end{document}